\documentclass[letterpaper, 10 pt, conference]{ieeeconf}  

\IEEEoverridecommandlockouts                              

\usepackage{amsfonts,amssymb,amsmath}
\usepackage{graphics}
\usepackage{epsfig}
\usepackage[all]{xy}
\usepackage{todonotes}
\usepackage{hyperref}

\usepackage{theorem}
\usepackage{morefloats}
\usepackage{amsfonts,amssymb,amsmath}
\usepackage{makeidx}

\def\hista{{\tilde{a}}} 
\def\histu{{\tilde{u}}} 
\def\histx{{\tilde{x}}} 

\def\sgn{{\rm sgn}}

\def\C{{\cal C}}

\def\Cfree{{\cal C}_{free}}

\def\qinit{{q_{I}}}
\def\qgoal{{q_{G}}}
\def\qdotinit{{{\dot q}_{I}}}
\def\qdotgoal{{{\dot q}_{G}}}

\def\Xfree{{X_{free}}}

\def\xinit{{x_{I}}}

\def\xgoal{{x_{G}}}

\def\x{{ x}}

\def\genseq{{\alpha}}

\def\Re{{\mathbb{R}}}

\def\qq1{{q_1}}

\def\qkp1{{q_{k+1}}}
\def\qKp1{{q_{F}}}

\def\qKp1{{q_{F}}}

\def\qxq1{{q_1}}

\def\uu1{{u_1}}

\def\xi{{x_i}}
\def\xkp1{{x_{k+1}}}
\def\xKp1{{x_{F}}}

\def\xq1{{x_1}}
\def\ukp1{{u_{k+1}}}

\def\argmin{\operatornamewithlimits{argmin}}

\def\atan2{\operatorname{atan2}}

\def\qdot{{\dot q}}

\def\qddot{{\ddot q}}

\def\xdot{{\dot x}}

\def\thetaddot{{\ddot \theta}}

\theoremstyle{plain}

\newtheorem{problem}{Problem}
\newtheorem{proposition}{Proposition}

\def\idz{{\hspace{0.25in}}}

\newcommand{\qed}{\hfill $\blacksquare$ \hfill \\}

\newcommand{\rdots}{\mathinner{
  \mkern1mu\raise1pt\hbox{.}
  \mkern2mu\raise4pt\hbox{.}
  \mkern2mu\raise7pt\vbox{\kern7pt\hbox{.}}\mkern1mu}}

\title{\LARGE \bf
Bang-Bang Boosting of RRTs
}

\author{Alexander J. LaValle, Basak Sakcak, and Steven M. LaValle
\thanks{The authors are with Center for Ubiquitous Computing, Faculty of Information Technology and Electrical Engineering,        University of Oulu, Finland.
       Email: {\tt\small firstname.lastname@oulu.fi}}%
}

\begin{document}

\maketitle
\thispagestyle{empty}
\pagestyle{empty}
\begin{abstract}

This paper presents methods for dramatically improving the performance of sampling-based kinodynamic planners.  The key component is the first-known complete, exact steering method that produces a time-optimal trajectory between any states for a vector of synchronized double integrators.  This method is applied in three ways: 1) to generate RRT edges that quickly solve the two-point boundary-value problems, 2) to produce a (quasi)metric for more accurate Voronoi bias in RRTs, and 3) to iteratively time-optimize a given collision-free trajectory.  Experiments are performed for state spaces with up to 2000 dimensions, resulting in improved computed trajectories and orders of magnitude computation time improvements over using ordinary metrics and constant controls.

\end{abstract}

\section{INTRODUCTION}

Rapidly exploring random trees were originally introduced as an approach to motion planning with differential constraints and dynamics \cite{LavKuf01b}.  The idea was to incrementally grow a space-filling tree by applying controls so that two-point boundary-value problems could be avoided if popular methods such as probabilistic roadmaps \cite{KavSveLatOve96} were applied to these problems.  Curiously, RRTs have found more success over the past decades for basic path planning (no differential constraints and dynamics), rather than their intended target, the {\em kinodynamic planning problem} \cite{DonXavCanRei93}.

Although this phenomenon is partly due to larger mainstream interest since the 1990s in basic path planning, it is primarily due to the additional challenges posed by the harder problems.  Computational performance depends greatly how well the RRT nearest-neighbor metric approximates the true, optimal cost-to-go function, which is presumably unattainable.  Furthermore, efficient steering methods or motion primitives are often needed to enhance performance, rather than applying constant controls as in \cite{LavKuf01b}.   Indeed, the most successful kinodynamic RRT planning methods have exploited the existence of simple cost-to-go functions and steering methods (ignoring obstacles) for special classes of systems \cite{FraDahFer05,GlaTed10,PerPlaKonKae12,SchJanPav15,WebVan13} or have relied on numerical solutions computed offline for more general systems and optimization objectives \cite{SakBasFerPra19}. 
Learning-based approaches to estimate the cost-to-go function have also been proposed \cite{ChiHsuFisTapFau19,LiBek11,PalArr15,WolBhaMoeWis18}.
Inspired by all of these works, we enhance RRT performance by using metrics and steering methods based on bang-bang time-optimal controls \cite{PonBolGamMis86}.  It was already shown that RRT exploration seems to improve with bang-bang metrics \cite{GlaTed10,KarFra10b}.

We continue in this direction by developing a steering method that completely solves the problem of time-optimally steering a vector of double integrators from any initial state to any goal state with a synchronized arrival time.  To the best of our knowledge, this is the first such solution to this problem, and optimal solutions are easily and exactly computed as two- to four-piece constant controls for each double integrator, resulting in piecewise-constant controls for the whole system, and trajectories as parabolic arcs in the configuration and state (phase) spaces.  The challenge is to ensure that all double integrators arrive at their goal states at the same time, which is often impossible due to momentum, unless some form of time-stretching or waiting is inserted.

We then show experimentally that the steering method improves RRT performance by orders of magnitude when compared to the original method that uses weighted Euclidean metrics and constant controls over fixed time intervals.  The study presented here is focused on double integrator dynamics, which lies at the core of fully actuated systems, with the intention of extending it to more general dynamics of arbitrary stabilizable systems (similar to the way it was accomplished in \cite{HeiJacCanPad90}).   As a step in this direction, we also present some preliminary results for a non-double integrator system, representing a vehicle on a curved surface.

This paper also presents methods that rapidly optimize collision-free trajectories by iteratively applying the simple time-optimal steering method to the output of sampling-based planners.  We consider two cases: 1) directly optimizing the result of a kinodynamic RRT-based planners, and 2) converting the piecewise-linear path produced by an RRT-based planner for basic path planning \cite{KufLav00} into a trajectory by applying bang-bang controls along each segment and then further iteratively optimizing the result.  Our experiments indicate that the second method is more efficient; however, it is limited to problems in which the initial and goal states are at zero velocity (if some form of completeness is demanded).
An alternative to these optimizations would be to apply asymptotically optimal extensions of RRTs, such as RRT* \cite{KarFra11} or SST* \cite{LiLitBek16}; however, we are motivated by the evidence that ``plan first and optimize later" often produces optimal paths more quickly and consistently than asymptotically optimal planning \cite{HeiPalKoeArrSuk18,LuoHau14}.

The paper continues as follows.  Section \ref{sec:problem} defines the problem.  Section \ref{sec:bb} develops the algebraic details of bang-bang time optimal control for single and multiple, parallel double integrators.  Section \ref{sec:plan} presents the new planning and optimization methods.  Section \ref{sec:exp} gives implementation details, computed results, and performance analysis.  Section \ref{sec:con} summarizes the results, their implications, and discusses the logical next steps.







\section{PROBLEM DEFINITION}\label{sec:problem}

Let $\C$ be the robot {\em configuration space}, assumed to be an $n$-dimensional smooth manifold with points referenced using local coordinates on $\Re^n$.  Geometric models (typically piecewise-linear) are given for the robot and its (static) environment, and the robot model transforms for each $q$ depending on robot kinematics.  Let $\Cfree$ be the open subset of $\C$ in which the robot does not intersect obstacles.  See \cite{ChoLynHutKanBurKavThr05,Lav06} for more details.  

The first problem is:

\begin{problem}[Basic path planning]
Given any $\qinit,\qgoal \in \Cfree$, compute a path $\tau : [0,1] \rightarrow \Cfree$ such that $\tau(0) = \qinit$ and $\tau(1) = \qgoal$.
\end{problem}
This problem ignores kinematic constraints and dynamics, leading to a second, harder problem that takes these into account.  Building upon $\C$ and $\Cfree$, let $x = (q,\qdot)$ be a $2n$-dimensional {\em state} vector for every configuration $q \in \C$.  The set of all $x$ forms $X$, the {\em state space}, which is the tangent bundle $T(\C)$.  Assume $\Cfree$ is lifted into $X$ as $\Xfree = \{ (q,\qdot) \in X \;|\; q \in \Cfree \}$.  

Let $\xdot = f(x,u)$ define a standard {\em control system} on $X$, in which $u$ belongs to a compact {\em action set} $U \subset \Re^m$.   For convenience in this paper, we will equivalently define the control system in terms of the accelerations that actions $u \in U$ induce at a particular $q \in \C$.  Thus, let $A(x) = A(q,\qdot)$ be the set of all accelerations $\qddot$ that can be obtained at $(q,\qdot)$ by applying an action $u \in U$ for the system $f$. Let $a \in A(x)$ denote a particular acceleration vector ($a = \qddot$) that may be applied. Let $A = \cup_{x \in X} A(x)$. 
Let $\Phi(x,\hista)$ denote the state trajectory $\histx:[0,t_F]\rightarrow X$ obtained by integrating a control $\hista:[0,t_F] \rightarrow A$ from state $x$.  This leads to the next problem:

\begin{problem}[Kinodynamic planning]
Given any $\xinit,\xgoal \in \Xfree$, compute an acceleration {\em control} $\hista : [0,t_F] \rightarrow A$ for which $\hista(t) \in A(\histx(t))$ for all $t \in [0,t_F]$, and that produces a state trajectory $\histx = \Phi(\xinit,\hista)$, with $\histx : [0,t_F] \rightarrow \Xfree$ and $\histx(t_F) = \xgoal$.
\end{problem}
We also consider {\em time optimality} in some cases, which means that a solution is chosen for which $t_F$ is as small as possible among all possible solutions. 

These restricted versions will be considered in this paper:
\begin{enumerate}
\item  {\em Stabilizable:} For all $x \in X$, $A(x)$ contains an open set that contains the origin $\bf 0$.  This implies that for any $\xinit$, $\xgoal$, there exists a finite-time acceleration control that solves the kinodynamic planning problem if $\Xfree = X$ (no obstacles).
\item {\em Rest-to-rest:} These problems require that for $\xinit=(\qinit,\qdotinit)$ and $\xgoal = (\qgoal,\qdotgoal)$, $\qdotinit = \qdotgoal = {\bf 0}$.
\item {\em nD-double-integrator:} Each $q_i$ is independently actuated within global acceleration bounds $a_{min,i} < 0$ and $a_{max,i} > 0$.  In this case, $A(x)$ is an axis-aligned rectangle that contains the origin, fixed for all $x \in X$.
\end{enumerate}


\section{TIME-OPTIMAL ACCELERATIONS}\label{sec:bb}



The task in this section is to calculate time-optimal solutions to the kinodynamic problem for the $n$D double integrator model and no obstacles: $\Xfree = X = \Re^{2n}$.  The solutions will work out so that controls are piecewise-constant,
\begin{equation}
\hista = ((a_1,t_1), (a_2,t_2), \ldots, (a_n,t_n)) ,
\end{equation}
which means that each $a_i$ is applied for duration $t_i$, starting at time $t_1 + t_2 \ldots t_{i-1}$.  Initially, $a_1$ is applied at time $t = 0$.

\subsection{Time-optimal control of a double integrator}\label{sec:bbo}

Consider one double integrator, for which every $q \in \Re$.  The allowable accelerations form a closed interval $A = [a_{min},a_{max}]$, in which $a_{min} < 0$ and $a_{max} > 0$.  It is a control system of the form $\qddot = a$ for $a \in [a_{min},a_{max}]$ and has an associated {\em phase plane}, with coordinates $(q,\qdot) \in \Re^2$.
The task is to determine an acceleration control $\hista:[0,t_F]\rightarrow A$ for which $\Phi(\xinit,\hista)=\xgoal$ and $t_F$ is as small as possible.

Pontryagin's maximum principle provides necessary conditions on the time-optimal trajectory by considering a co-state vector $(\lambda_1,\lambda_2)$ that serves as a generalized Lagrange multipliers for the constrained optimization problem \cite{Lib12}.  Following the standard theory, the Hamiltonian is defined as
\begin{equation}\label{eqn:hamdefdi}
H(x,a,\lambda) = 1 + \lambda_1 x_2 + \lambda_2 a ,
\end{equation}
in which $x_1 = q$ and $x_2 = \qdot$, and the optimal $\hista$ is constrained to
\begin{equation}\label{eqn:minhamdi}
\hista^*(t) = \argmin_{a \in A} \left\{ 1 + \lambda_1(t) x_2(t) + \lambda_2(t) \hista(t)\right\}.
\end{equation}
Solving the adjoint equation ${\dot \lambda_i} = -\frac{\partial H}{\partial x_i}$ results in $\lambda_1(t) = c_1$ and $\lambda_2(t) = c_2 - c_1t$ for unknown constants $c_1$ and $c_2$.
If $\lambda_2(t) < 0$, then $\hista^*(t) = a_{max}$, and if $\lambda_2(t) > 0$,
then $\hista^*(t) = a_{min}$.  Thus, the action may be assigned as $\hista^*(t) =
-\sgn(\lambda_2(t))$, if $\lambda_2(t) \not = 0$.  At the boundary
case in which $\lambda_2(t) = 0$, any $a \in A$ may be
chosen.  Since $\lambda_2(t)$ is linear, it may change signs at most once, implying that the optimal control involves at most two ``bangs", each corresponding to an extremal acceleration applied over a bounded time interval.  Thus, the time-optimal control is of the form  $\hista^* = ((a_1,t_1),(a_2,t_2))$, with degenerate possibilities of $t_1=0$ or $t_2=0$.

\begin{figure}
\centerline{\psfig{figure=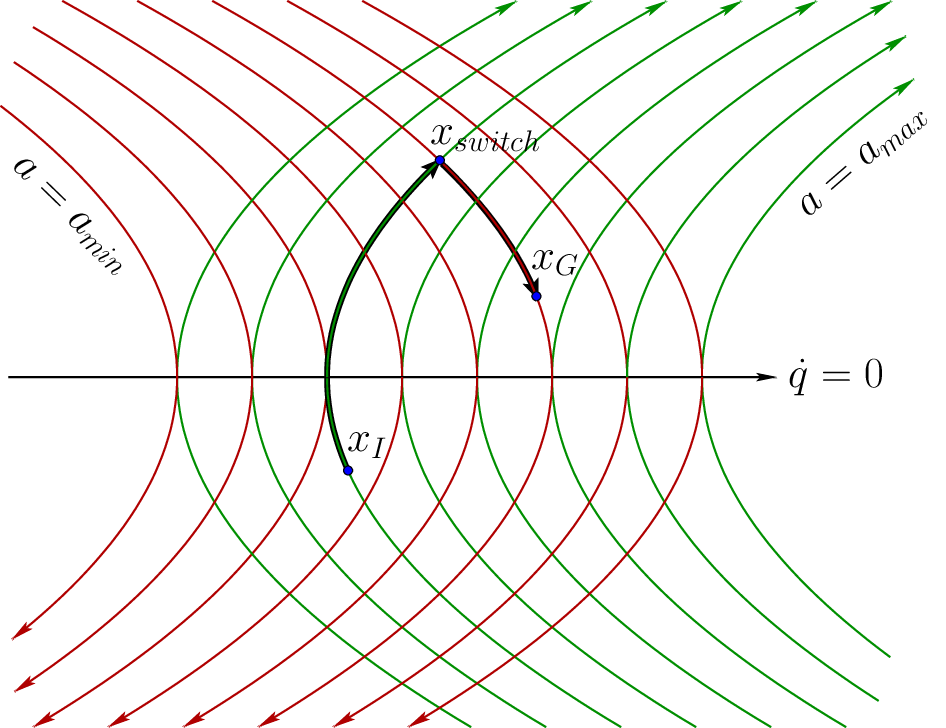,width=7cm}}
\caption{\label{fig:phaseparabolas} Intersecting parabolas and traveling forward in time produces the time-optimal trajectory in the phase plane, which generally involves two-piece constant controls and parabolic trajectories.}
\end{figure}

Further algebraic analysis is needed to precisely determine $a_1$, $t_1$, $a_2$, and $t_2$ for a given $(\qinit,\qdotinit)$ and $(\qgoal,\qdotgoal)$.  See Figure \ref{fig:phaseparabolas}.  If a constant acceleration $a$ is applied at any phase $(q_0,\qdot_0)$, then a parabolic curve is traced out in the phase plane.  If $q_0 =\qdot_0= 0$, then its equation is $q = \frac{1}{2}\xdot^2$.  More generally, it satisfies $q-\frac{1}{2}\qdot^2 = c$, in which the coefficient $c = q_0 - \frac{1}{2}\qdot_0^2$ corresponds to the parabola's intersection with the $\qdot=0$ axis.

Next consider four parabolas based on all combinations of initial and goal states, and extremal controls. Let $I^+$ denote the parabola obtained from setting $(q_0,\qdot_0) = \xinit = (\qinit,\qdotinit)$ and applying constant control $a_{max}$; let $c(I^+)$ denote its $\qdot=0$ intercept.  Similarly, $I^-$ is obtained by applying $a_{min}$.  Furthermore, $G^+$ and $G^-$ are obtained by setting $(q_0,\qdot_0) = \xgoal$ and applying $a_{max}$ and $a_{min}$, respectively.  Assuming $\xinit \not = \xgoal$, consider all possible intersections of $I^+$, $I^-$, $G^+$ and $G^-$.  There are only two possible types: $I^+G^-$ and $I^-G^+$.  The first results in $a_{max}$ applied until the intersection point, $x_{switch}=(q_{switch},\qdot_{switch})$, is reached, followed by applying $a_{min}$.  The second type applies $a_{min}$ first, followed by $a_{max}$.

The type $I^+G^-$ intersection occurs if $c(I^+) > c(G^-)$.  The intersection position is the midpoint $q_{switch} = (c(I^+) + c(G^-))/2$, and the velocity is $\qdot_{switch} = \sqrt{2(q_{switch}-c(I^+))}$.  Similarly, the type $I^-G^+$ intersection occurs if $c(I^-) < c(G^+)$.  The intersection position is the midpoint $q_{switch} = (c(I^-) + c(G^+))/2$, and the velocity is $\qdot_{switch} = -\sqrt{2(q_{switch}-c(G^+))}$.

To determine the control timings to go from $\xinit$ to $x_{switch}$ to $\xgoal$ note that changing velocity by an amount $d$ with constant acceleration $a$ requires time $d/a$.  The timings are:
\begin{equation}\label{eqn:t1}
t_1 = (\qdot_{switch} - \qdotinit) / a_1
\end{equation}
and
\begin{equation}\label{eqn:t2}
t_2 = (\qdotinit - \qdot_{switch}) / a_2,
\end{equation}
respectively.

There will always be at least one intersection, but sometimes there are both $I^+G^-$ and $I^-G^+$.  In this case, $t_1$ or $t_2$ may be negative for one intersection, but the other intersection provides a valid control.  It is also possible to obtain valid controls for both cases, in which case the one that requires least time must be selected (Figure \ref{fig:wait} will provide more details).

The arguments above lead to the following proposition:
\begin{proposition}\label{prop:dio}
The time-optimal control for the double integrator problem is $((a_1,t_1),(a_2,t_2))$, in which $a_1 = a_{min}$ and $a_2 = a_{max}$, or $a_1 = a_{max}$ and $a_2 = a_{min}$, and the durations $t_1$ and $t_2$ are given by (\ref{eqn:t1}) and (\ref{eqn:t2}).  
\end{proposition}


\subsection{Time-optimal control of a vector of double integrators}\label{sec:waste}

To extend the result to a vector of double integrators, a waiting method is needed so that each double integrator arrives at its goal in its phase plane at the same time.  This problem was considered in \cite{FraDahFer02,KarFra10b}, but only for the limited case in which their final velocities are zero.  It was also considered in \cite{HauNgt10}, but the gap problem, introduced shortly, was neglected.  In this section, we introduce an explicit, complete, and computationally efficient solution to the general problem of time-optimal steering of $n$ independent integrators for any initial and goal state pairs in their respective phase planes in $O(n \lg n)$ time.

\begin{figure}
\centerline{\psfig{figure=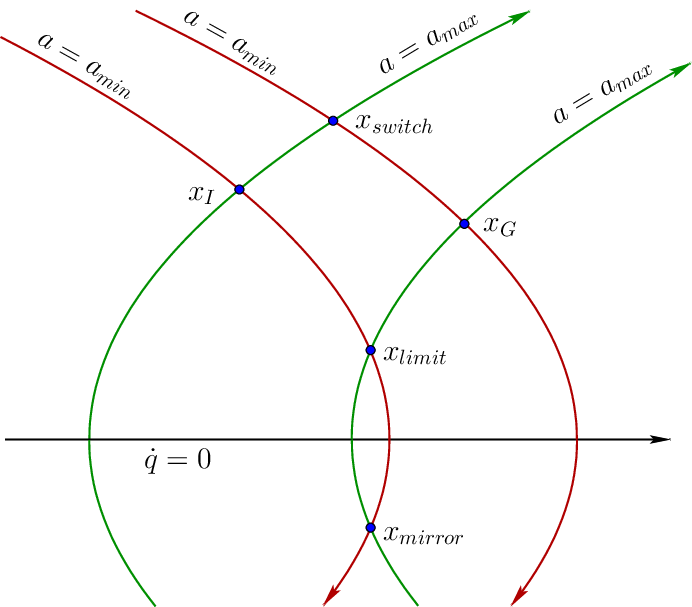,width=7cm}}
\caption{\label{fig:wait} If both $I^+G^-$ and $I^-G^+$ intersections occur, then there is a gap interval, which disallows certain waiting times.  This poses a significant challenge to synchronizing the arrival times of $n$ double integrators, which is overcome in our paper and is critical to our new planning algorithms.}
\end{figure}

For a fixed $\xinit$ and $\xgoal$, let $t^*$ be the time to reach the goal by applying the bang-bang solution of Section \ref{sec:bbo}.  Now consider some $t_w > t^*$.  Does a control $\histu$ necessarily exist that will cause $\xgoal$ to be reached at exactly time $t_w$?  The answer is yes if there is only one intersection type ($I^+G^-$ or $I^-G^+$).  However, if both intersections occur, then the situation depicted in Figure \ref{fig:wait} occurs (or its symmetric equivalent for $\qdot < 0$).  The path from $\xinit$ to $x_{limit} = (q_{limit},\qdot_{limit})$ to $\xgoal$ corresponds to the slowest trajectory that reaches $\xgoal$ while remaining in the $\qdot > 0$ half-plane.  The critical switching point $x_{limit}$ can be calculated using the parabola intersection algebra from Section \ref{sec:bbo}.  The time taken by this trajectory is calculated as
\begin{equation}
t_{limit} = (\qdot_{limit}-\qdotinit)/a_{min} + (\qdotgoal-\qdot_{limit})/a_{max}.
\end{equation}
Thus, if $t_w \in [t^*,t_{limit}]$, then a solution exists (and requires only two constant control segments).  If $t_w > t_{limit}$, then $\xgoal$ can no longer be reached while remaining in the $\qdot > 0$ half-plane.  The next available time is obtained by continuing to apply $a = a_{min}$ until the second parabolic intersection point, called $x_{mirror}$ is reached, in the $\qdot < 0$ half-plane.  Note that $\x_{mirror} = (q_{limit},-\qdot_{limit})$.  Once $x_{mirror}$ is reached, $a = a_{max}$ is applied to arrive optimally at $\xgoal$.  The time required to traverse this trajectory is 
\begin{equation}
t_{mirror} = t_{limit} + 2 \qdot_{limit} / a_{max} -  2 \qdot_{limit} / a_{min}.
\end{equation}
Thus, there may generally be a {\em gap interval} $(t_{limit},t_{mirror})$ for which no solution exists.  

Now suppose that $t_w \geq t^*$ and there is no gap interval, or $t_w \in [t^*,t_{limit}] \cup [t_{mirror},\infty)$.  A two-piece control $((a_1,t_1),(a_2,t_2))$ can be calculated as a solution to the following equations corresponding to the boundary conditions:
\begin{equation}\label{eqn:bc_pos}
    \qdotinit t_1 + \dfrac{a_1t_1^2}{2} + (\qdotinit+a_1t_1)t_2 + \dfrac{a_2t_2^2}{2}=\qgoal-\qinit,
\end{equation}
\begin{equation}\label{eqn:bc_vel}
    a_1t_1 + a_2t_2=\qdotgoal-\qdotinit,
\end{equation}
while satisfying $a_1,a_2\in [a_{min},a_{max}]$.
Note that $t_2=t_w-t_1$ and $0 \leq t_1,t_2\leq t_w$. The solution is usually not unique, and can be selected either arbitrarily or by optimizing a relevant optimization objective, such as energy. 

\begin{proposition}
If a solution exists for prescribed time $t_w$ to reach $\xgoal$ from $\xinit$, then the above waiting method generates a control that achieves it; otherwise, it reports failure (implying that $t_w$ lies in the gap interval).
\end{proposition}

\noindent {\bf Proof:} 
Let $\tilde{a}_u$ and $\tilde{a}_l$ be two control trajectories defined as $\tilde{a}_u = ((a_{max}, t_u), (a_{min}, t_w-t_u))$ and $\tilde{a}_l=((a_{min}, t_l), (a_{max},t_w-t_l))$. Furthermore, for given $\xinit$ and $\xgoal$, $t_u$ and $t_l$ satisfy that $\qdot_u(t_w)={\qdot_l}(t_w)=\qdotgoal$, in which $(q_u,\qdot_u)=\Phi(\xinit, \tilde{a}_u)$ and $(q_l,\qdot_l)=\Phi(\xinit, \tilde{a}_l)$. Suppose a solution exists for $t_w$ and let $\Tilde{a} : [0, t_w] \rightarrow [a_{min}, a_{max}]$ be a solution control trajectory such that $\Tilde{x}(t_w)=\xgoal$, in which $\Tilde{x}=(q, \qdot) = \Phi(\xinit, \Tilde{a})$. For all $t \in [0, t_w]$, it is true that 
\begin{equation}\label{eqn:prop2_vel_bounds}
\qdot_l(t) \leq \qdot(t) \leq \qdot_u(t).
\end{equation}
Suppose this is not true and there exists a $t' \in [0, t_w]$ for which $\qdot(t') > \qdot_u(t')$. Then, if $t' \leq t_u$ it implies that $\int_0^{t'} \Tilde{a}(t)dt > \int_0^{t'} \Tilde{a}_u(t)dt$ which violates the condition that $\Tilde{a}(t) \leq a_{max}$ for all $t \in [0, t_w]$. The same reasoning can be done for $t'>t_u$ and for $\qdot(t) < \qdot_l(t)$. It follows from \eqref{eqn:prop2_vel_bounds} that 
\begin{equation}\label{eqn:prop2_pos_bounds}
q_l(t) \leq q(t) \leq q_u(t)
\end{equation}
for all $t \in [0, t_w]$. In particular, $q_l(t_w) -\qinit \leq \qgoal-\qinit \leq q_u(t_w) -\qinit$. Then, there exist $a_1,a_2 \in [a_{min}, a_{max}]$ and $0\leq t_1 <t_w$ that satisfy the boundary conditions given in \eqref{eqn:bc_pos} and \eqref{eqn:bc_vel}. This proves that if a solution $\hista$ exists for $t_w$, then, there also exists a two-piece solution $((a_1, t_1), (a_2, t_w-t_1))$. As a consequence, if 
there does not exist a two-piece solution, then, there also does not exist a solution for $t_w$. 
This happens for example, if $\qgoal < q_l(t_w)$ or $\qgoal > q_u(t_w)$ for given $t_w, \xinit$, and $\xgoal$. 
\qed

Putting these results together, a control $\histu$ can be determined for any $t_w$, unless it is in the gap interval.  If $t_w \geq t_{mirror}$ and $q_{init} \geq 0$, then a four-piece solution is obtained by: 1) maximum deceleration to rest from $\xinit$, 2) waiting for time $t_w - t_{mirror}$, 3) maximum deceleration to $x_{mirror}$, and 4) maximum acceleration to $\xgoal$.  A symmetric equivalent solution applies for $q_{init} \leq 0$.  If $t_w \in [t^*,t_{limit}]$, then the two-piece solution from (\ref{eqn:bc_pos}) and (\ref{eqn:bc_vel}) is used (this method could even be applied if $t_w \geq t_{mirror}$, but this was not attempted).

Now suppose that the optimal times have been calculated for $n$ double integrators to start at some $\xinit$ and end at some $\xgoal$.  In the worst case, every double integrator could have a gap interval.  The problem is to find the smallest time $t$ such that $t \geq t^*$ and $t \not \in (t_{limit},t_{mirror})$.  If a double integrator has no gap interval, then that part of the condition is dropped.  A simple algorithm is contained in the proof of the following proposition:


\begin{proposition}
The time-optimal steering problem for $n$ double integrators can be solved in $O(n \lg n)$ time.
\end{proposition}

\noindent {\bf Proof:}  A simple line sweeping algorithm \cite{DebVanOveSch97} achieves the bound as follows.  Sort the optimal, limit, and mirror times for all double integrators into a single array of length $O(n)$.  Sweep incrementally across the array, starting from the shortest time.  In each step increment or decrement a counter of the number of double integrators that have a solution; $t_{limit}$ causes decrementing and the other two cases cause incrementing.  Each step takes $O(1)$ time.  The method terminates with the optimal $t$ when all $n$ integrators admit a solution.  The overall algorithm takes time $O(n\lg n)$ time due to the initial sorting. \qed



\section{PLANNING METHODS}\label{sec:plan}

\subsection{Kinodynamic RRT with bang-bang metric and steering}\label{ssec:bbmetric}

\begin{figure}
\noindent \rule{\columnwidth}{0.25mm}
BANG\_BANG\_RRT\_BIDIRECTIONAL($\xinit,\xgoal$) \\
\begin{tabular}{ll}
1 & $T_a$.init($\xinit$); $T_b$.init($\xgoal$) \\
2 & {\bf for} $i = 1$ {\bf to} $K$ {\bf do} \\
3 & \idz $x_n \leftarrow ${\sc bang-bang-nearest}($S_a,\genseq(i)$) \\
4 & \idz $x_s \leftarrow ${\sc bang-bang-steer}($x_n$,$\alpha(i)$) \\
5 & \idz {\bf if} $x_s \not = x_n$ {\bf then} \\
6 & \idz \idz $T_a$.add\_vertex($x_s$) \\
7 & \idz \idz $T_a$.add\_edge($x_n,x_s$) \\
8 & \idz \idz $x^\prime_n \leftarrow $ {\sc bang-bang-nearest}($S_b,x_s$) \\
9 & \idz \idz $x^\prime_s \leftarrow $
              {\sc bang-bang-steer}($x^\prime_n$,$x_s$) \\
10 & \idz \idz {\bf if} $x^\prime_s \not = x^\prime_n$ {\bf then} \\
11 & \idz \idz \idz $T_b$.add\_vertex($x^\prime_s$) \\
12 & \idz \idz \idz $T_b$.add\_edge($x^\prime_n,x^\prime_s$) \\
13 & \idz \idz {\bf if} $x^\prime_s = x_s$ {\bf then return}
       SOLUTION \\
14 & \idz {\bf if} $|T_b| > |T_a|$ {\bf then} SWAP$(T_a,T_b)$ \\
15 & {\bf return} FAILURE \\
\end{tabular} \\
\rule{\columnwidth}{0.25mm}
\caption{\label{fig:birrt} Bidirectional RRT with bang-bang steering and quasimetric.  }
\end{figure}

Suppose a kinodynamic planning problem is given for an $n$D-double integrator system.  Figure \ref{fig:birrt} presents an outline of a balanced bidirectional RRT-based planning algorithm that uses bang-bang methods for both the metric and the steering method.  A single-tree {\em goal-bias} algorithm could alternatively be made \cite{LavKuf01b}.  Let $\alpha(i) \in X$ denote the random state obtained in iteration $i$ ($\alpha$ could alternatively be a deterministic sequence that is dense in $X$ \cite{Lav06}).  Line 3 returns the nearest state $x_n$ among all points $S_a$ visited by tree $T_a$.  Using the tools from Section \ref{sec:bb}, there are two natural choices for the (quasi)metric, $\rho(x,x')$, which is an estimate of the distance from $x$ to $x'$.  Note it is not symmetric for our problem.  The first choice $\rho_1(x,x')$ is the maximum time $t_1+t_2$ from (\ref{eqn:t1}) and (\ref{eqn:t2}), taken over all $n$ double integrators.  A slower and more accurate metric is $\rho_2(x,x')$ is the time with waiting, $t_w$, from Section \ref{sec:waste}, which is the time it takes for every double integrator to arrive at $x'$.  Note that these metrics are expected to produce a better Voronoi-bias \cite{LavKuf01b} because they are closer to the true optimal cost-to-go function.


Line 4 is the time-optimal steering method from Section \ref{sec:waste}.  Collision checking is performed along the trajectory, and the steering stops at $x_s$ if an obstacle is hit or $\alpha(i)$ is reached.  The new trajectory is added to $T_a$ (and $S_a$).  In practice, this was accomplished in our experiments by inserting into $T_a$ nodes and edges along the trajectory; an exact method could alternatively be developed for representing and computing nearest points in $S_a$. 

Lines 8 and 9 are similar to Lines 3 and 4, except that an attempt is made to connect the newest visited point $x_s$ to the nearest point $S_b$ in the other tree, $T_b$.  Line 14 swaps the roles of the tree so that the smaller one explores toward $\alpha(i)$ and the larger one attempts to connect to the newly reached point.

If a more general, stabilizable system (recall from Section \ref{sec:problem}) is given, then it can be converted into an $n$D-double integrator by restricting $A$ to an compact, axis-aligned rectangular region that contains the origin.   Such a subset of $A$ always exists, and lies in the intersection of the open subsets of $A(x)$ that contain ${\bf 0}$, for all $x \in X$.  If the rectangle is small relative to $A(x)$ at each $x$, then we expect the solutions produced by the algorithm in Figure \ref{fig:birrt} to be further from their potential optima; a step toward investigating this problem is taken at the end of Section \ref{sec:exp}.

\subsection{Bang-bang trajectory optimization}\label{sec:bbto}

Let $X$, $\Xfree$, $\xinit$, $\xgoal$, and $A$ be fixed for an $n$D double integrator system.  Suppose that a piecewise-constant control $\hista:[0,t_F]\rightarrow A$ is given so that the resulting $\histx = \Phi(\xinit,\hista)$ is solution to Problem 2 from Section \ref{sec:problem}.  The task is to replace $\hista$ with a new control $\hista':[0, t_F'] \rightarrow A$ so that $t'_F < t_F$ while maintaining the constraints that the trajectory maps into $\Xfree$ and arrives at $\xgoal$ at time $t'_F$.  The new control $\hista'$ is constructed by selecting $t_1$ and $t_2$ such that $0 \leq t_1 < t_2 \leq t_F$.  

For a given control, $\hista$, let $\hista[t_1,t_2]$ denote its restriction to the interval $[t_1,t_2]$.  Thus, $\hista[t_1,t_2] : [t_1,t_2] \rightarrow A$.\footnote{The restrictions will be closed intervals that allow single-point overlaps, but this will not affect the resulting trajectories.}  The original control $\hista$ can be expressed as a sequence of three controls $\hista[0,t_1]$, $\hista[t_1,t_2]$, and $\hista[t_2,t_F]$ by applying $\Phi$ to each in succession.  We replace the middle portion with a bang-bang control $\hista'[t_1,t'_2]$ using the methods of Section \ref{sec:waste}.  Since the method is time-optimal, it is known that $t'_2 \leq t_2$ (they are equal only if $\hista[t_2,t_F]$ is already time-optimal).  The new control must satisfy
\begin{multline}
\Phi(\Phi(\xinit,\hista[0,t_1])(t_1),\hista[t_1,t_2])(t_2) \\ =
\Phi(\Phi(\xinit,\hista[0,t_1])(t_1),\hista'[t_1,t_2'])(t_2') ,   
\end{multline}
which implies that the new control sequence $\hista[0,t_1]$, $\hista[t_1,t'_2]$, and $\hista[t_2,t_F]$ arrives at $\xgoal$ at time $t_f - (t_2 - t'_2)$. Fresh collision checking is needed for the trajectory from $\Phi(\xinit,\hista[0,t_1])(t_1)$ to $\Phi(\Phi(\xinit,\hista[0,t_1]),\hista'[t_1,t_2'])(t_2')$.

The general template for {\em iterative bang-bang optimization} is:
\begin{enumerate}
\item Choose $t_1$ and $t_2$ according to a random or deterministic rule.
\item Attempt to replace $\hista[t_1,t_2]$ with the bang-bang alternative $\hista'[t_1,t'_2]$.  If the result is collision free, then update $\hista$ with the modified control. 
\item Go to Step 1, unless a termination criterion is met based on the number of iterations without any significant time reduction.
\end{enumerate}
The rule of choosing $t_1$ and $t_2$ should produce a dense sequence of intervals in the following sense: the points of the form $(t_1,t_2) \in \Re^2$ must be dense in the triangular region satisfying $0 \leq t_1 \leq t_2 \leq t_F$ (this ignores the fact that $t_F$ decreases in each iteration, and such out-of-bounds intervals can be rejected in the analysis).  A simple but effective rule is to first pick $t_1$ and $t_2$ uniformly at random.  If $t_1 < t_2$, then replace $\hista[t_1,t_2]$.  Otherwise, toss an unbiased coin to replace either $\hista[0,t_2]$ or $\hista[t_1,t_F]$.  This extra consideration over purely random pairs (e.g., as in \cite{HauNgt10}) helps focus on the ends.  Alternatively, $t_1$ and $t_2$ could be picked according to deterministic sequences to ensure convergence.

The termination criterion could be based on a hard limit on the number of iterations, or failure statistics (for example, no significant improvement more than $\epsilon > 0$ has occurred in the past 50 iterations).  Note that the approach is not a variational optimization as in a gradient descent in trajectory space \cite{Bet98}; it more resembles path shortening for basic path planning (called {\em shortcutting} in \cite{GerOve07,LuoHau14}).  Thus, local time-optimality is gradually reached in the sense that the solution cannot be further improved by replacing trajectory segments with time-optimal alternatives, but it is not equivalent to a time optimum in the sense of local perturbations in trajectory space.

\subsection{Basic path planning with bang-bang state-space lifting}\label{sec:lift}

Consider taking the output of a basic path planning, lifting it into the state space using bang-bang control, and then applying the bang-bang optimization method of Section \ref{sec:bbto}.  Suppose we are given a kinodynamic planning problem for $n$D double integrators for which $\xinit$ and $\xgoal$ are both at rest (zero velocity).  Thus, $\xinit = (\qinit,{\bf 0})$ and $\xgoal = (\qgoal,{\bf 0})$.  The first step is to compute a piecewise-linear path $\tau: [0,1] \rightarrow \Cfree$, in which $\Cfree$ is the projection of $\Xfree$ onto the configuration space.  This could, for example, be computed by RRT-Connect \cite{KufLav00}, but the particular planner is unimportant.

We then introduce the {\em bang-bang transform}, described here for $\Re^n$ and $a_{max} = -a_{min}=1$ (it easily generalizes; see also \cite{HauNgt10}).  For each vertex $q$ along the path $\tau$, extend it to $x=(q,{\bf 0}) \in \Xfree$.  For each edge between consecutive vertices, $q$, $q'$, execute a bang-bang control; we require that it is constrained to the edge and steers from $(q,{\bf 0})$ to $(q',{\bf 0})$.  Let $v = q' - q$, normalized as $\hat{v} = v / \|v\|$.  Let $s = \max_i(|\hat{v}_i|)$.  Let $a_i = \hat{v}_i / s$ and $t = \sqrt{s\|v\|}$.  The bang-bang control is $((a,t),(-a,t))$.  This transform is applied to each edge of the path and the resulting controls are concatenated.  

The following propositions support the approach of lifting any piecewise-linear collision-free path (which are the typical output of sampling-based planners) into the state space via the bang-bang transform.

\begin{proposition}\label{prop:bbt}
The bang-bang transform of a path is time-optimal.  Furthermore, $\Phi(\xinit,\hista) = \xgoal$ and the resulting trajectory is collision free.
\end{proposition}

\noindent {\bf Proof:}  Assuming that there are no spurious vertices (lying in the interior of a linear segment), the system must come to rest at each vertex.  Thus, applying the time-optimal control from rest to rest over each edge yields a time-optimal control for the whole path.  The bang-bang transform is merely a consequence of Proposition \ref{prop:dio}, applied to the simpler rest-to-rest case.  Regarding collision, for each segment, the accelerations yield velocities parallel to it; thus, the path traversed is the edge itself, which is already known to be collision free.  Furthermore, this implies that $\xgoal$ is reached after the full control $\hista$ is applied.
\qed

\begin{proposition}
Using the bang-bang transform, any piecewise-linear solution to Problem 1 can be converted via the bang-bang transform into a corresponding solution to Problem 2, restricted to double integrators and rest-to-rest.
\end{proposition}

\noindent {\bf Proof:}  This is a direct consequence of Proposition \ref{prop:bbt} due to the preservation of the collision-free and goal reachability properties of the bang-bang transform.
\qed




\section{EXPERIMENTS}\label{sec:exp}

The algorithms were implemented in Python 3.9.5 on a Windows 10 PC with an AMD Ryzen 7 5800X CPU and 32GB 3200MHz CL16 RAM.  Naive methods were used for nearest neighbor searching and collision detection because they are not critical to the experimental analysis.  All results are shown in a high-resolution video available at \url{http://lavalle.pl/videos/IROS23.mp4}.

\subsection{Kinodynamic planning for a 2D vehicle (4D state space)}\label{sec:4d}

These examples use a four-dimensional state space corresponding to a 2D workspace in which a planar vehicle moves with double integrator dynamics.  Let $a_{max} = -a_{min} = 1$.

\begin{figure}\label{fig:res1}
\begin{tabular}{cc}
\psfig{figure=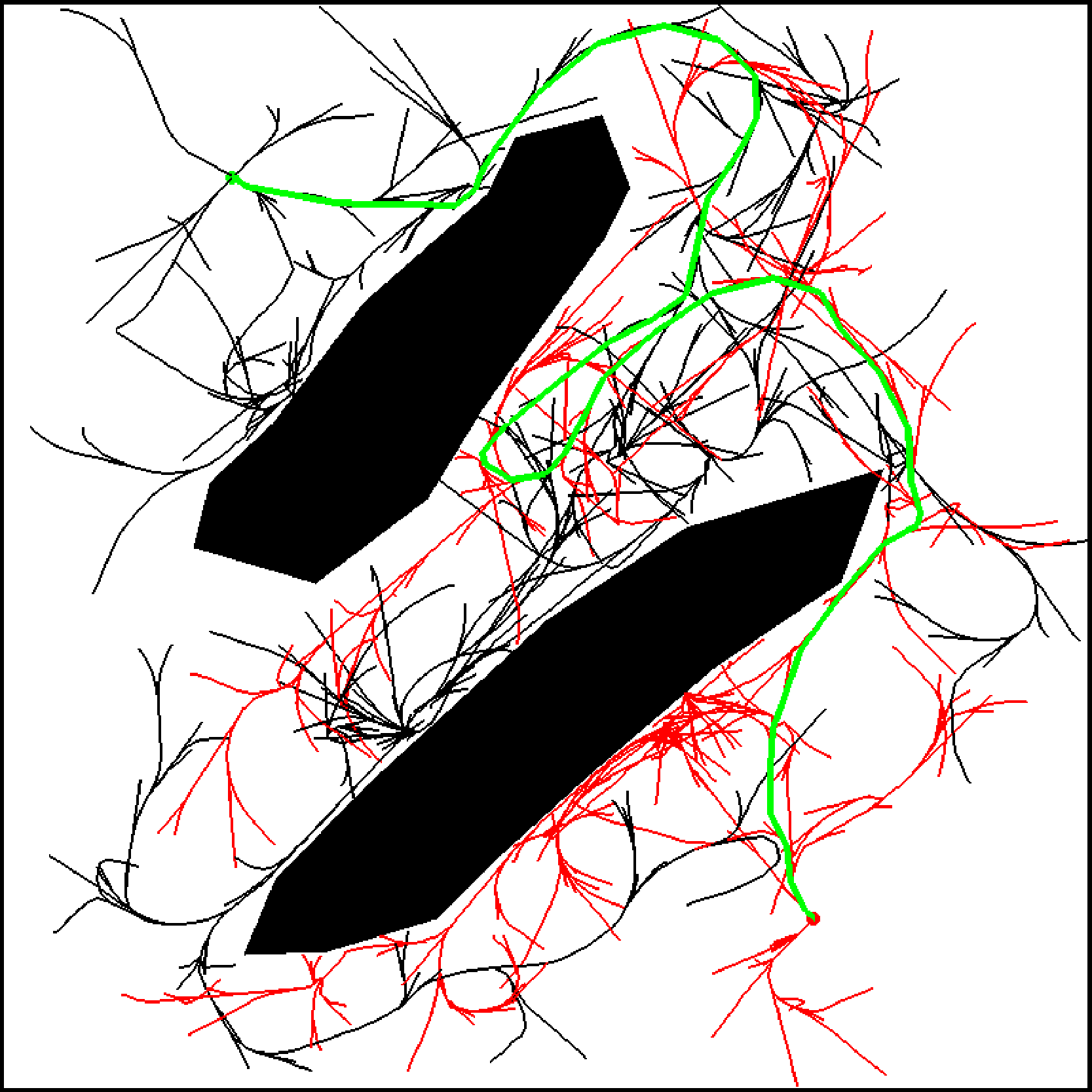,width=4cm} & \psfig{figure=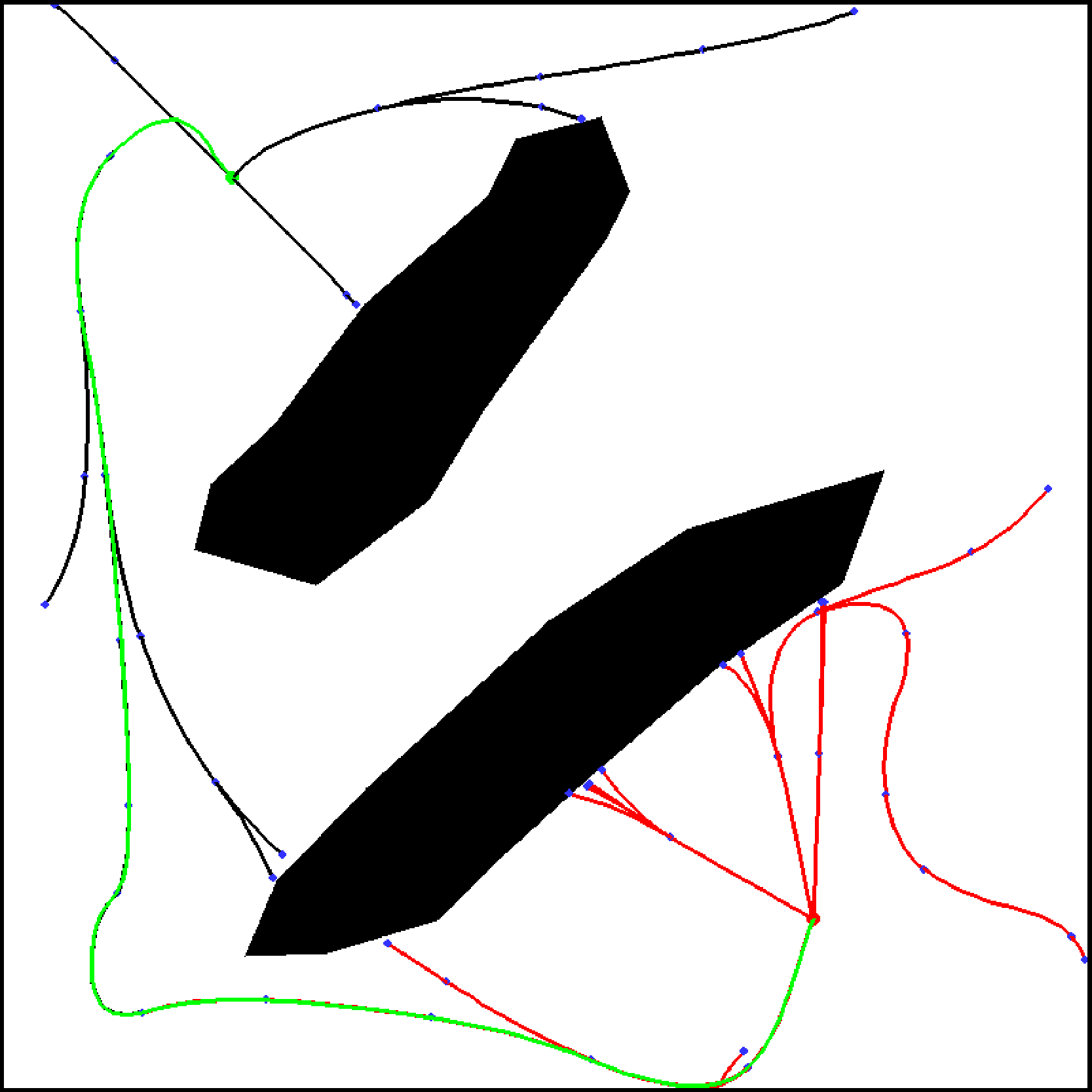,width=4cm} \\
a. & b. \\
\psfig{figure=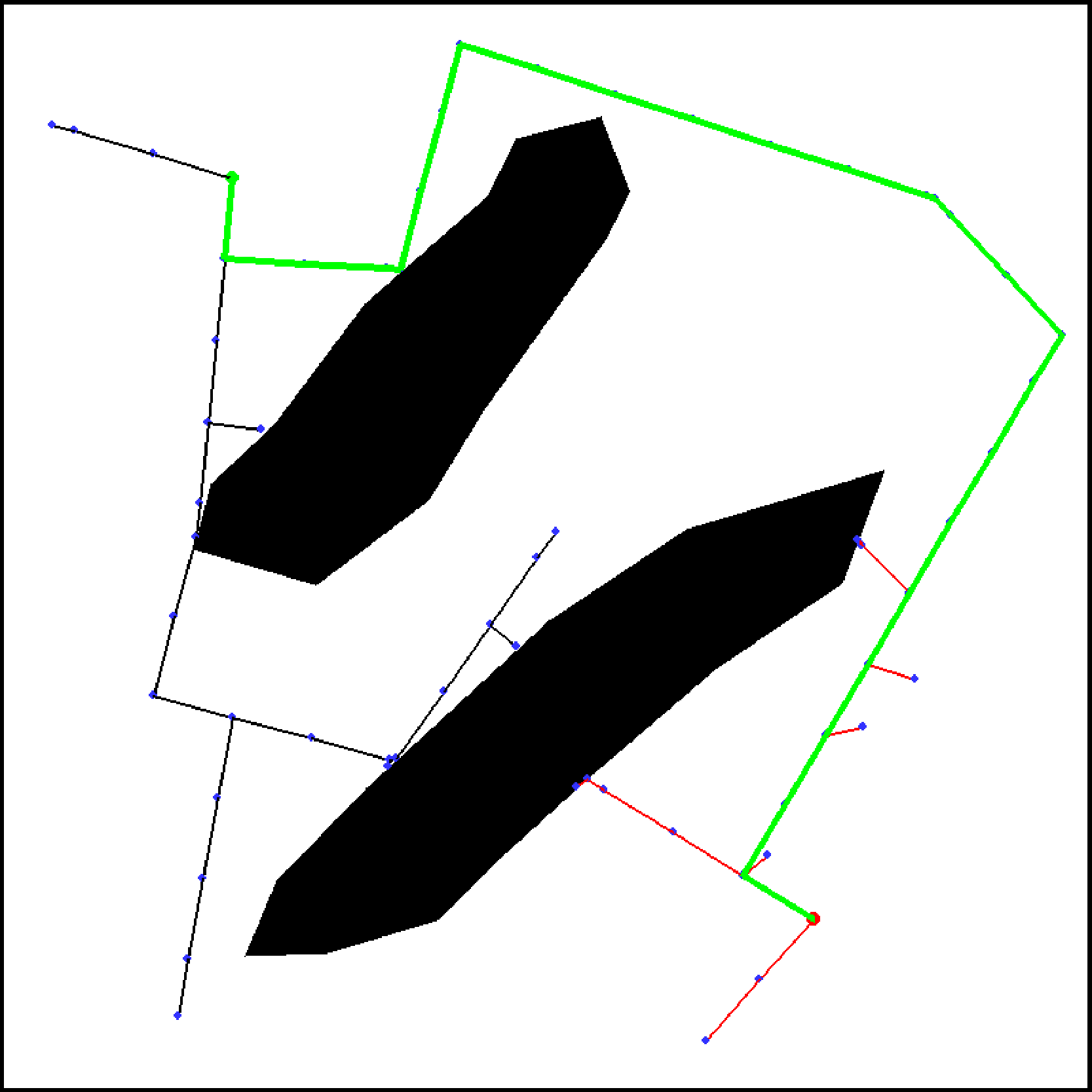,width=4cm} & \psfig{figure=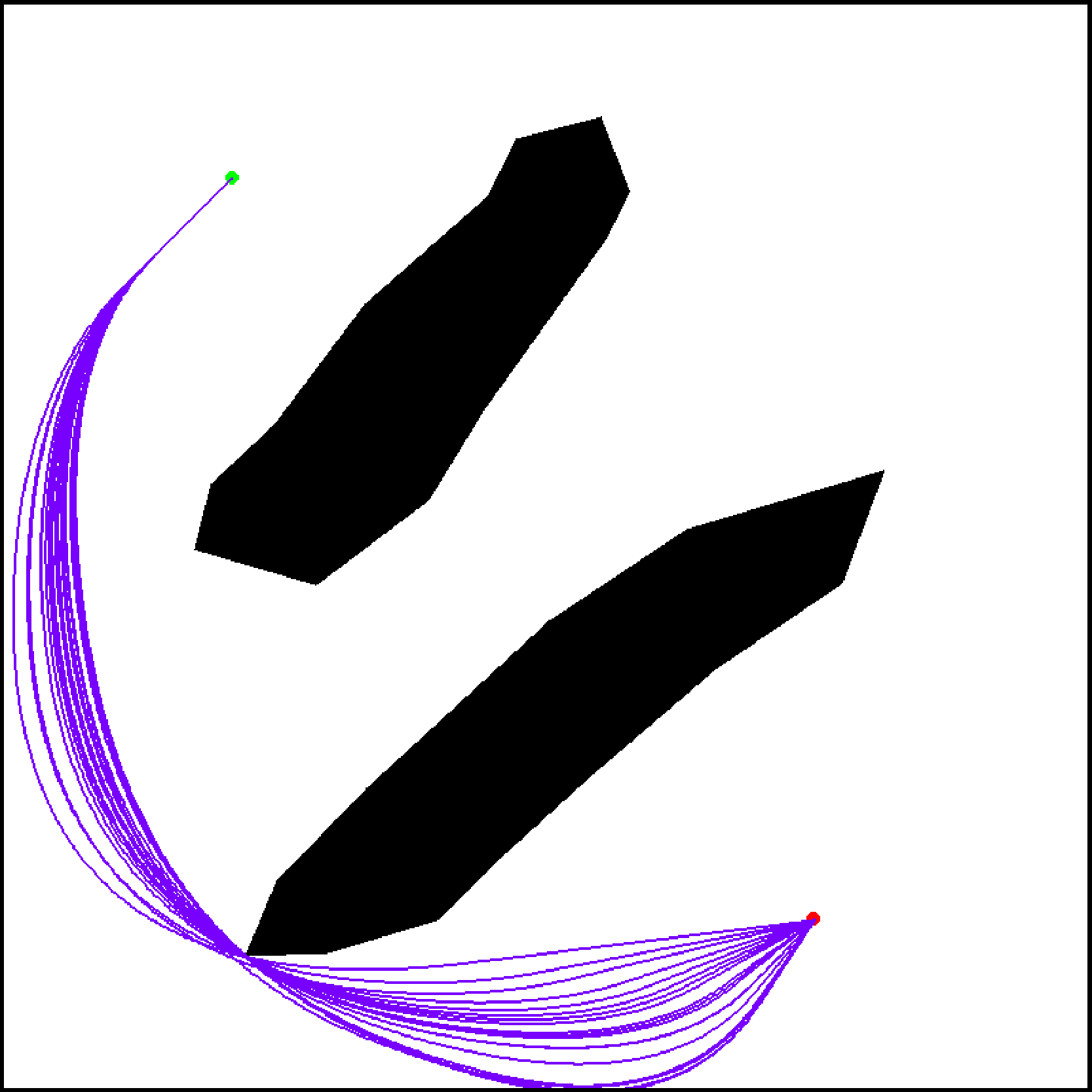,width=4cm} \\
c. & d. \\
\psfig{figure=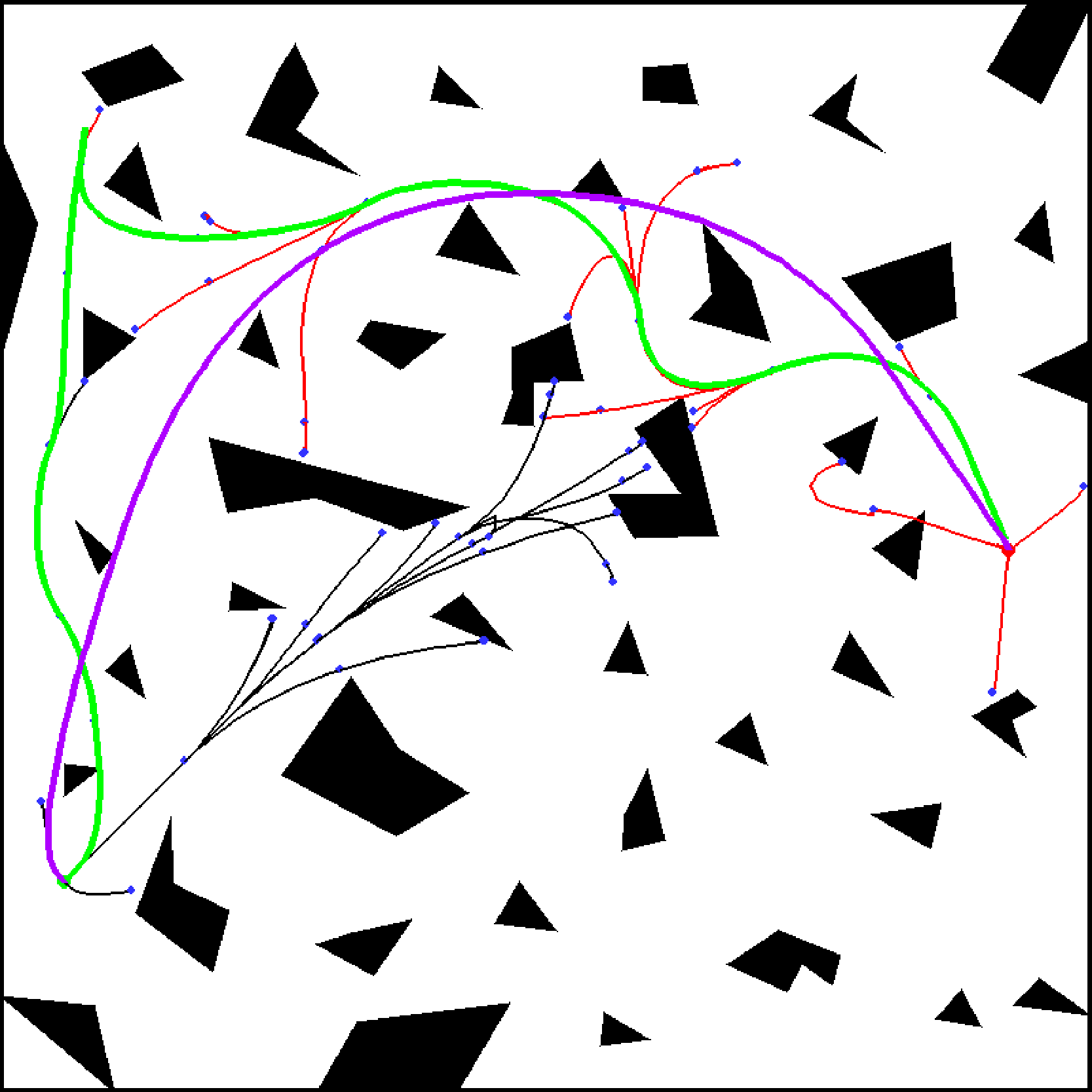,width=4cm} & \psfig{figure=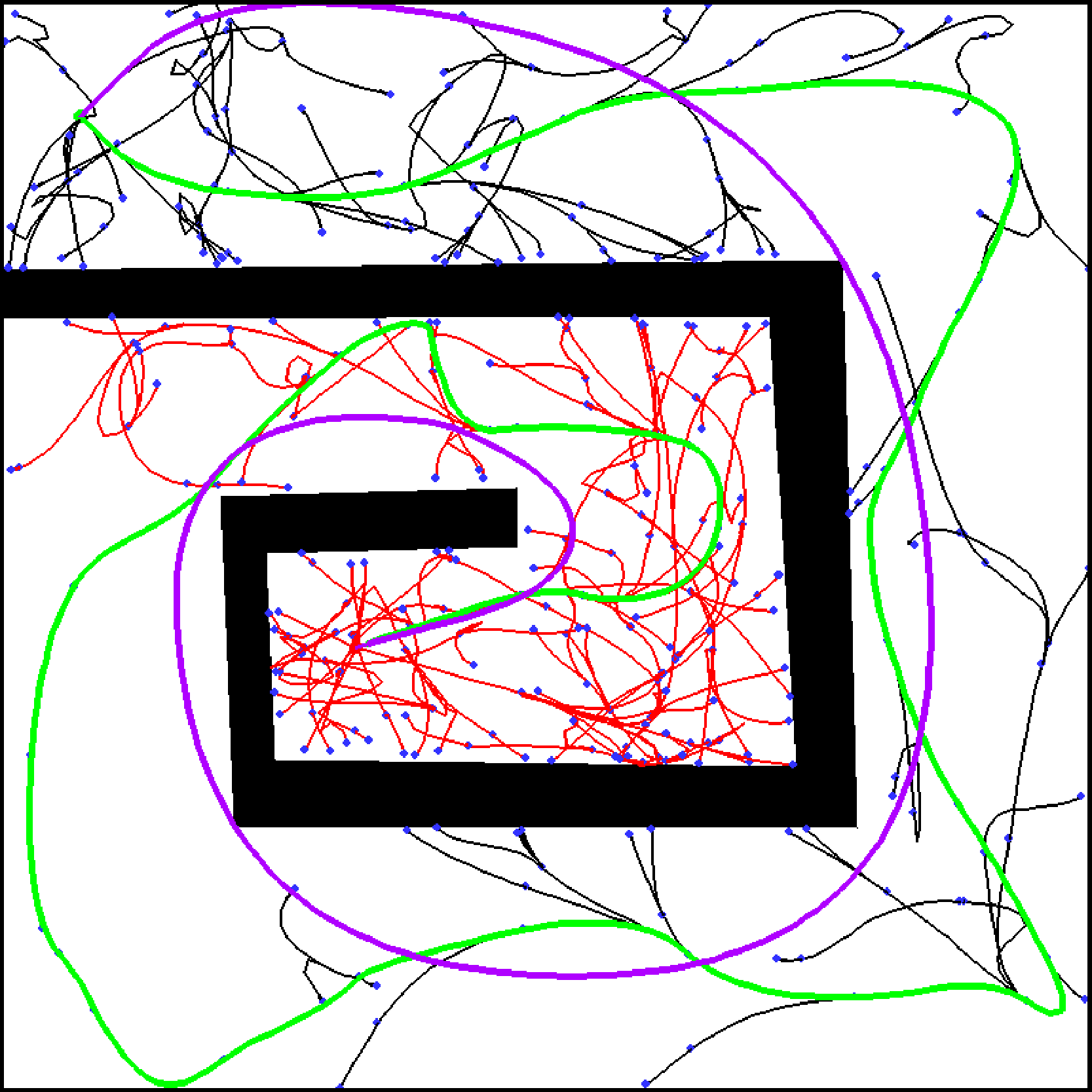,width=4cm} \\
e. & f.
\end{tabular}
\caption{\label{fig:rrt} a) Original kinodynamic RRT-Bi \cite{LavKuf01b}, b) The proposed BB-RRT, c) RRT-Connect \cite{KufLav00} applied to the 2D projection, d) multiple bang-bang optimizations of a planned path, e \& f) two more BB-RRT examples with BB-optimized paths (purple).}
\end{figure}




\begin{figure}
\begin{tabular}{|l|l|l|l|l|}\hline
Method & RunTime & Nodes & ColChecks & TrajTime \\ \hline
RRT-Bi & 27.013 & 1589.1 & 4364.8 & 311.53 \\ \hline
BB-RRT & 0.017276 & 57.772 & 599.48 & 126.99 \\ \hline
RRT-Con & 0.004738 & 60.663 & 713.25 & n/a \\ \hline
BB-Opt & 0.021547 & n/a & 3291.3 & 72.452 \\ \hline
\end{tabular}
\caption{\label{fig:stats} For each method, the execution time (second), number of RRT nodes, number of collision checks, and trajectory execution time (seconds) are reported (where applicable).  All numbers are calculated as averages over 1000 runs.}
\end{figure}

Figure \ref{fig:rrt} compares the new BB-RRT method (Figure \ref{fig:birrt}) to the original kinodynamic RRT-Bi \cite{LavKuf01b} and RRT-Connect \cite{KufLav00} on the 2D projection that ignores velocities and dynamics.  The state space $X$ is $[-400,400]^2 \times [-10,10]^2$. Statistics are reported in Figure \ref{fig:stats}.   The initial and goal states were at rest.  The $\rho_1$ metric introduced in Section~\ref{ssec:bbmetric} was used.  The $S_a$ and $S_b$ sets were approximated by placing new RRT nodes along long edges for every 12 collision checks (see Section 5.5.2 of \cite{Lav06}).  The BB-RRT is about 1564 times faster on average than the RRT-Bi.  RRT-Connect is even faster, but it only constructs paths on the 2D configuration space, and the number of collision checks is comparable.  Figure \ref{fig:rrt}.d shows 50 results for the bang-bang optimizer of Section \ref{sec:bbto}, applied to the same initial path; computation times are fairly consistent across runs and problems, depending mainly on path length and collision detector cost.

The BB-RRT has the advantage, much like RRT-Connect, in that there are no parameters to tune.  RRT-Bi has parameters for the step size, the set of actions, and the connection distance (the trees do not exactly meet).  For the example in Figure \ref{fig:rrt}.a, we used 24 constant acceleration actions, $\Delta t = 5$, and connection distances of $\Delta q = 5$ and $\Delta \qdot = 2$;  in the weighted-Euclidean metric, the velocity components were weighted 17.32 times more than the configuration components.  Figures \ref{fig:rrt}.e and \ref{fig:rrt}.f show two more examples under the same conditions, for which BB-RRT took on average 0.0368s and 0.4072s, respectively, over 1000 runs.   The speedup factors over RRT-Bi were 837.6 and 368.5.  Again, RRT-Connect on the 2D projection was faster, by factors 6.65 and 12.5, respectively.  Original and bang-bang optimized paths are shown green and purple, respectively.



\begin{figure}
\begin{tabular}{cc}
\psfig{figure=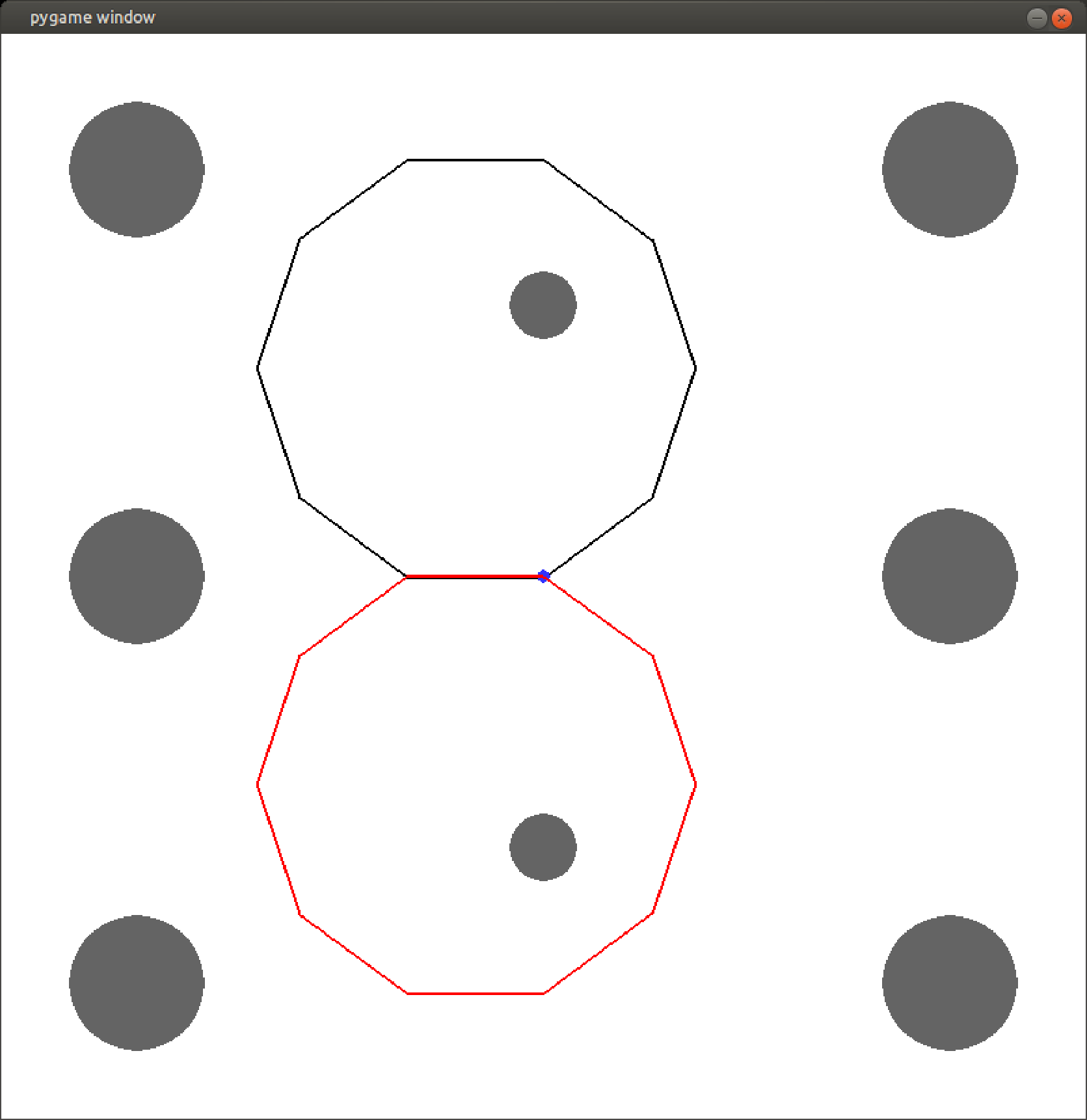,width=4cm} & \psfig{figure=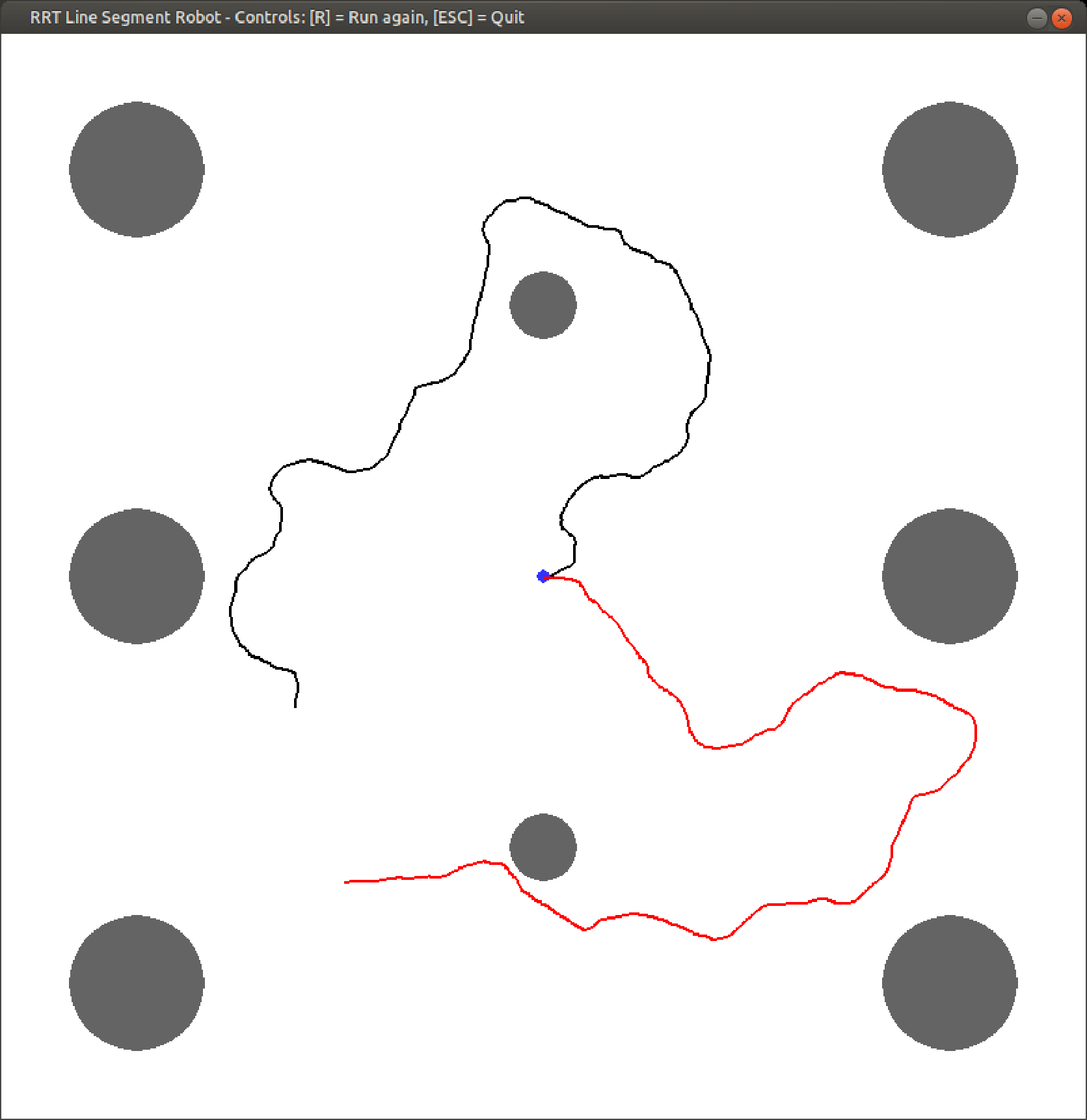,width=4cm} \\
a. & b.
\end{tabular}
\caption{\label{fig:lsr}  BB-optimization applied to a planar manipulator with dynamics.}
\end{figure}

\subsection{Bang-bang optimization for a planar manipulator}

Figure \ref{fig:lsr} depicts additional experiments, performed for an $n$-link, fixed-base planar manipulator, modeled as a kinematic chain of line segments of equal length.  Again, assume $a_{max} = -a_{min} = 1$.  The initial and goal configurations form a regular polygon, as shown in Figure \ref{fig:lsr}.a for $n=10$ links.  Paths were initially computed using RRT-Connect on $\C$ and then lifted into $X$ using the bang-bang transform of Section \ref{sec:lift}.  Each joint has limits $\pm \pi$ and is modeled as a double integrator.  The BB-Optimization method was applied to computed paths from $n=10$ (dimension of $X$ is 20) up to $n=1000$ (dimension of $X$ is 2000).  Figure \ref{fig:lsr}.b shows configurations in the RRTs that were grown from initial and goal configurations, respectively.  
Average running times (10 runs) to fully converge are $1.63$s for 10 links, $1.72$s for 20 links, $3.75$s for 50 links, $20.75$s for 100 links, and $1123.35$s for 1000 links (rapid increases due to dimension were caused by a naive quadratic-time implementation of the waiting method, rather than the $O(n\lg n)$ method presented in Section \ref{sec:waste}).  Termination was reached if 200 iterations were attempted with no more than $0.1$s reduction in trajectory time; the most dramatic reductions occur in the first few iterations.  In a typical run for 100 links, the trajectory execution time was reduced from $57.95$s to $8.46$s. 

\subsection{Beyond pure double integrator dynamics}\label{sec:ndi}

\begin{figure}
\begin{tabular}{cc}
\psfig{figure=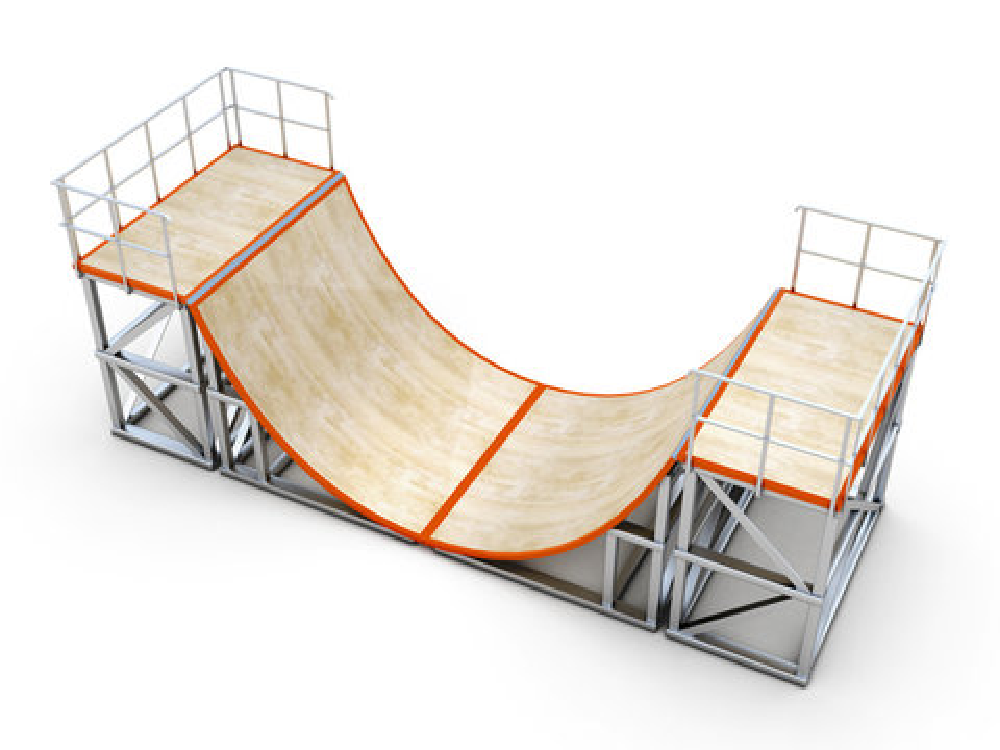,width=4cm} & \psfig{figure=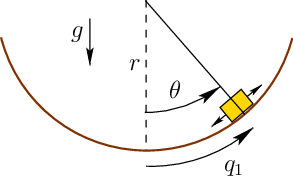,width=4cm} \\
a. & b. \\
\psfig{figure=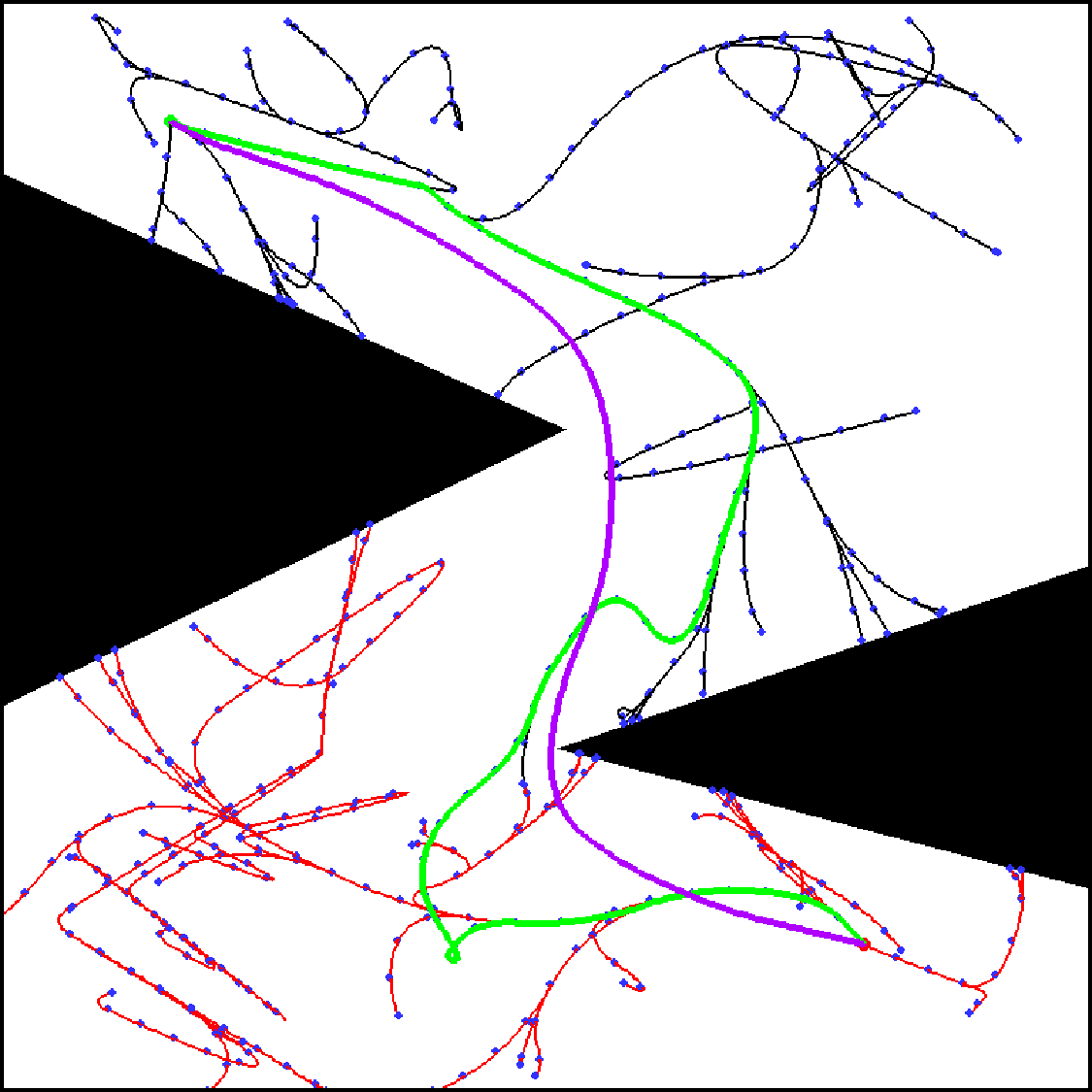,width=4cm} & \psfig{figure=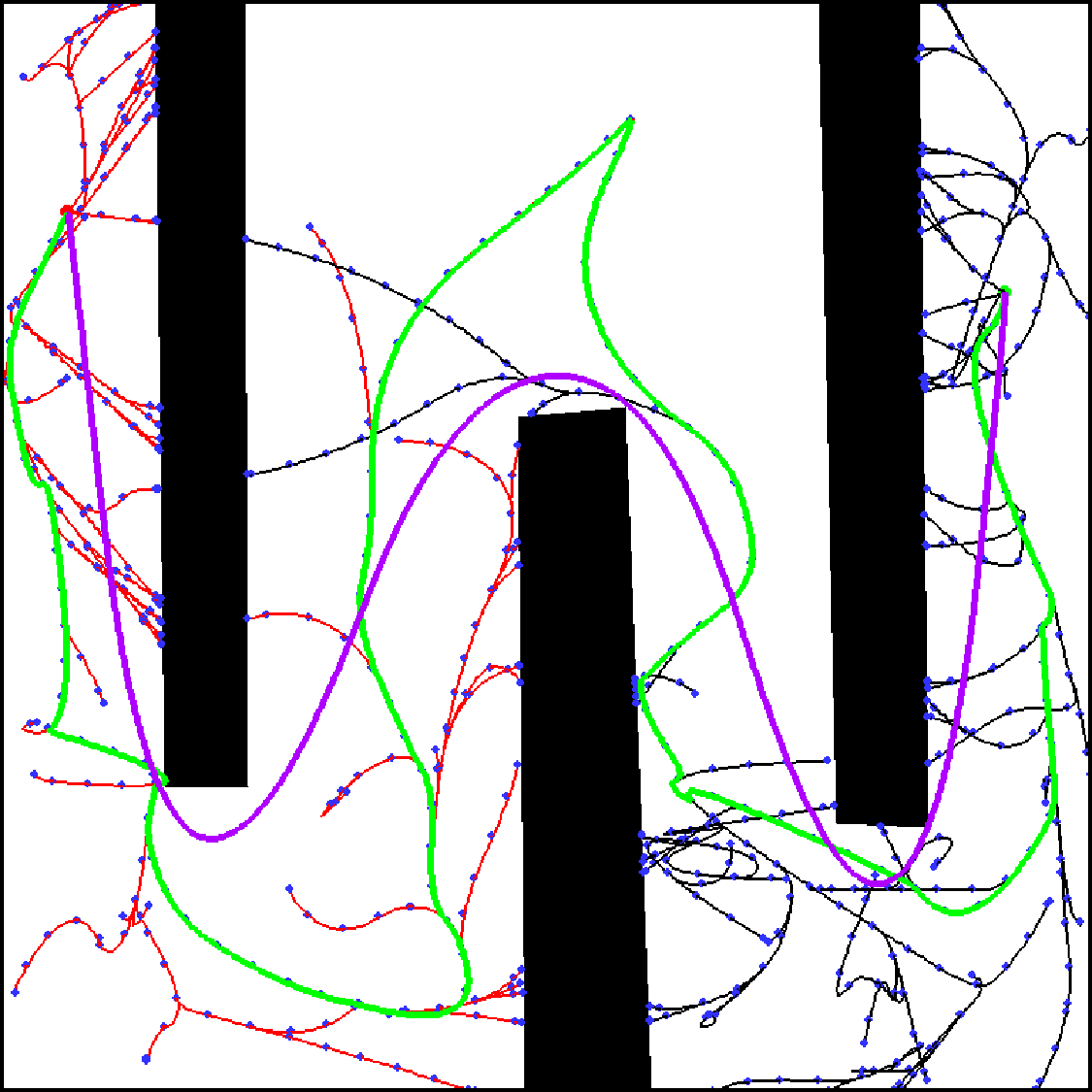,width=4cm} \\
c. & d. 
\end{tabular}
\caption{\label{fig:tube}  a) Consider a moving vehicle on a cylindrical surface, b) the $q_1$ coordinate behaves like an actuated pendulum, c \& d) computed examples (horizontal and vertical axes correspond to $q_1$ and $q_2$, respectively).}
\end{figure}

As a step toward investigating bang-bang boosting of more general, stabilizable systems, suppose that the planar vehicle from Section \ref{sec:4d} is instead placed on the interior surface of a level, cylindrical tube of radius $r$ (Figures \ref{fig:tube}.a-b).  The $q_1$ coordinate dynamics resemble that of an actuated pendulum:
\begin{equation}
\qddot_1 = r \thetaddot = u_1 - g \sin\theta .
\end{equation}
The $q_2$ coordinate is the position along the tube in the direction of its central axis, with dynamics $\qddot_2 = u_2$.  Assume $u_1,u_2 \in [-1,1]$.  If $|g\sin\theta| < 1$ for $\theta$, then the system is stabilizable, as defined in Section \ref{sec:problem}.  The interval of allowable accelerations $\qddot_1$ becomes $A(\theta) = [-1-g\sin\theta,1-g\sin\theta]$.  To generate a bang-bang trajectory from some $\theta_I$ to $\theta_G$, we restrict the system to a double integrator in which $a_{min}$ and $a_{max}$ are set to the minimum and maximum of $A(\theta_I) \cap A(\theta_G)$.  We also test and reject any generated bang-bang trajectory for which $\qddot_1 \not \in A(\theta)$ at any time. 

Two computed examples of both BB-RRT planning and bang-bang optimization are shown in Figures \ref{fig:tube}.c-d, in which the vehicle must go from rest-to-rest along the curved surface; a top-down view is given by unrolling the cylinder.  The state space $X$ is the same as in Section \ref{sec:4d}, $r = 300$, and $g = 1$; note that the slope $\theta = q_1/r$ along the left and right edges reaches $\pm 4/3$ radians ($76.394$ degrees).  The allowable horizontal accelerations at these boundaries are approximately $[-0.028,1.972]$ and $[-1.972,0.028]$, respectively (substantially shifted from $[-1,1]$).  The computation times averaged over 1000 runs were $0.02963$s and $1.1184$s, respectively.   We also ran 1000 experiments on the geometry of the problem in Figure 4.e, but instead using the tube model, and the resulting average computation time was $0.09594$s (approximately $2.61$ times slower than for the level-surface case).  In general, the planning and optimization methods easily overcame the challenges due to the non-double integrator model.

\section{DISCUSSION}\label{sec:con}

We have proposed, analyzed, and implemented methods that accelerate planning performance and optimize solutions.  The key is our new steering method that quickly computes bang-bang time-optimal controls using exact, parabolic solutions.  Although the study has been limited to RRTs, we expect it could enhance other sampling-based planning methods that rely on distance metrics or steering, such as {\em probabilistic roadmaps} \cite{AmaBayDalJonVal98,KavSveLatOve96} or {\em expansive space trees} \cite{HsuLatMot99}.  One of the key observations of our experiments is that plan-and-optimize is superior when applicable:  It is more reliable to explore the C-space first, lift the solution into the state space, and then use bang-bang optimization.  However, this option applies only for rest-to-rest problems; for more general problems, a bang-bang enhanced RRT could be applied to bring each of $\xinit$ and $\xgoal$ to zero velocity by biasing samples to the $(q,{\bf 0})$ plane.

The encouraging results of this paper lead naturally to many new questions and further studies.  The implementation focused mainly on $n$-double-integrator dynamics; however, with the vehicle-in-the-tube results from Section \ref{sec:ndi}, we have easily extended it for acceleration bounds that vary with state.  This opens exciting directions of research to adapt the method to many more classes of stabilizable systems.  Another important direction is to develop bang-bang boosted versions of asymptotically optimal planners, such as RRT* \cite{KarFra11} and SST* \cite{LiLitBek16}; this would enable stronger comparisons to the plan-and-optimize approach, both in computation time and solution quality.  Also, improvements can be made to the iterative bang-bang optimization through strategic interval selection.  Finally, efficient nearest-neighbor algorithms should be developed for the bang-bang metric over a tree of parabolic arcs (analogous to \cite{AtrLav02,ValPadYerFra16}).


\noindent \subsubsection*{ACKNOWLEDGMENTS}


{\small This work was supported by a European Research Council Advanced Grant (ERC AdG, ILLUSIVE: Foundations of Perception Engineering, 101020977), Academy of Finland (PERCEPT 322637, CHiMP 342556), and Business Finland (HUMOR 3656/31/2019).  We thank Dmitry Berenson, Kalle Timperi, and Dmitry Yershov for helpful discussions.}



\bibliographystyle{plain}
\bibliography{main,pub,new}


\end{document}